%% file: cvpr.tex
\documentclass[final]{cvpr}

\usepackage[utf8]{inputenc} 
\usepackage[T1]{fontenc}    
\usepackage{url}            
\usepackage{booktabs}       
\usepackage{amsfonts}       
\usepackage{nicefrac}       
\usepackage{microtype}      
\usepackage{times}
\usepackage{epsfig}
\usepackage{graphicx}
\usepackage{amsmath}
\usepackage{amssymb,bm}
\usepackage{cuted}
\usepackage{xcolor}
\usepackage{multirow}
\usepackage{caption}
\usepackage{subcaption}

\usepackage[T1]{fontenc}
\usepackage[font=small,labelfont=bf,tableposition=top]{caption}

\input{macros}

\usepackage[pagebackref=true,breaklinks=true,colorlinks,bookmarks=false]{hyperref}

\def\cvprPaperID{8652} 
\def\confYear{CVPR 2021}

\begin{document}

\title{RP2K: A Large-Scale Retail Product Dataset for Fine-Grained Image Classification}

\author{Jingtian Peng\\
Pinlan Data Technology Co,. Ltd.\\
\and
Chang Xiao\\
Columbia University\\
\and
Yifan Li\\
Pinlan Data Technology Co,. Ltd.\\

}

\maketitle

\input{abs}
\input{intro}

\input{related}
\input{org}
\input{baseline}

\input{conclusion}


{\small
\bibliographystyle{ieee_fullname}
\bibliography{ref.bib}
}

\newpage

\end{document}

%% file: macros.tex
\makeatletter
\renewcommand{\paragraph}{%
  \@startsection{paragraph}{4}%
  {\z@}{0.60ex \@plus 1ex \@minus .15ex}{-1em}%
  {\normalfont\normalsize\bfseries}%
}
\makeatother

%% file: abs.tex
\begin{abstract}
  We introduce RP2K, a new large-scale retail product dataset for fine-grained image classification.
  Unlike previous datasets focusing on relatively few products, we collect 350,000 images of more than 2000 different retail products, directly captured on shelves in real retail stores.
Our dataset aims to advance the research in retail object recognition, which has massive applications such as automatic shelf auditing and image-based product information retrieval. Our experiments show that even state-of-the-art fine-grained classification methods did not outperform a simple ResNet baseline, indicating a large potential space of research to improve the classification performance on the task of fine-grained retail product classification.
Our dataset enjoys the following properties: (1) It is by far the largest dataset in terms of product categories. (2) All images are captured manually in physical retail stores with natural lightings, matching the scenario of real applications. (3) We provide rich annotations to each object, including the sizes, shapes and flavors/scents. 
We believe our dataset could benefit both computer vision research and retail industry.
\end{abstract}

%% file: intro.tex
\section{Introduction}

\begin{figure}[t]
  \centering
    \includegraphics[width=1.0\columnwidth]{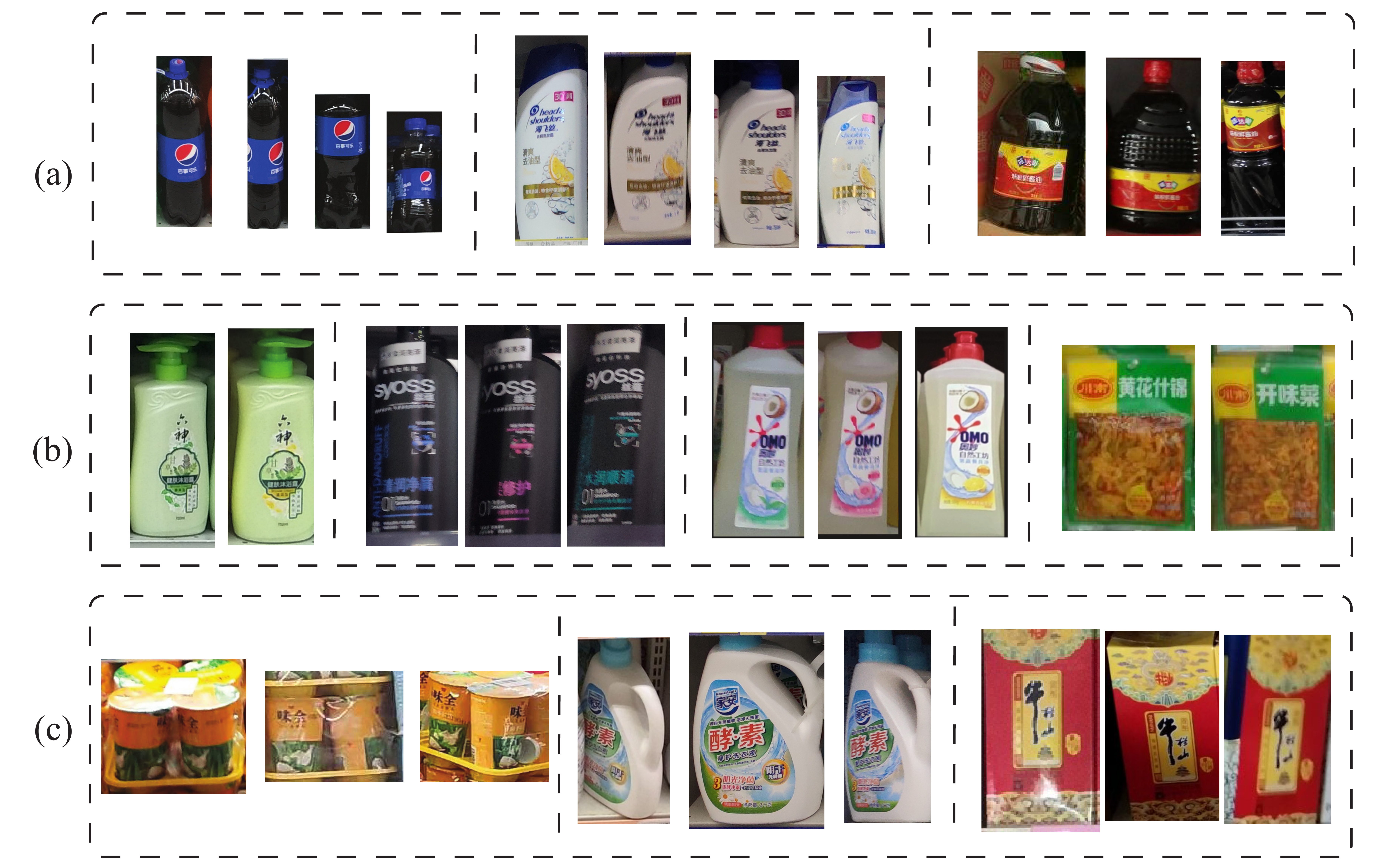}
    \caption{Sample images from our dataset. Precise retail product recognition on shelves is considered highly challenging because \textbf{(a)} Products from the same line may have different sizes, and they usually have similar appearances but different prices. The image size could not reflect the real size of the products. \textbf{(b)} The manufacturer usually make multiple flavors for one product line, but their appearance only have subtle differences on the labels. \textbf{(c)} Product images may be captured at different camera angles according to its placement location on shelves. The image can also be stretched due to camera distortion.
  }\label{fig:example}
  \vspace{-6mm}
\end{figure}

AI is reshaping the retail industry.
The traditional retail industry requires tremendous human labor for its entire supply chain, from product distribution to inventory counting.
In response to the rise of online shopping, traditional markets also quickly take advantage of AI-related technology at the physical store level to replace the tedious human labor by computer algorithms and robots. 
Many promising applications of applying AI to retailing are considered to be widely realized in the near-term. 
For instance, by using computer vision techniques, the retailers can audit the placement of products or track in-store sale activities from shelf images, 
and the customers can obtain the product information by taking pictures of the objects.
To implement these applications, one of the core problem is to develop a robust recognition system that can recognize the products placed on shelves.

Despite the recent advances in computer vision, the task of recognizing retail products on shelf, is still considered to be challenging in the computer vision perspective.
First, the number of product categories can be huge in a supermarket. According to Goldman et al.~\cite{goldman2019precise}, 
a typical supermarket could have more than thousands of different products.
Second, different products may have similar appearances. 
For example, products from the same brand with different sizes or flavors usually have a similar appearance, but they should be recognized as separate products (see \figref{example}-a,b).
Third, the camera angle and lighting condition of the product to be recognized may vary a lot (see \figref{example}-c), which requires a reliable recognition algorithm that can handle these scene complexities. 
Moreover, many applications, like product auditing, requires the system to have a very high recognition accuracy. 
Even a small error of product recognition could lead to a large price difference, which may result in unpredictable amount of loss for the retailer.
That is to say, an algorithm that have 95\% accuracy may considered as a ``good'' computer vision algorithm, but it is still far from being practical to deployed to real retail activities to replace the human labor. We will discuss the detailed applications for applying AI in retail industry in \secref{app}.

To overcome these challenges, several retail product datasets have been proposed in the past decades ~\cite{wei2019rpc,goldman2019precise,grozi-120,follmann2018mvtec,hao2019take,supermarket}. However, most of the previous datasets either focus on a relatively small scale (less than 200 categories), or collect data in a controlled environment (e.g., lab with sufficient light).
To bridge the gap between research and real-life application, we propose the RP2K dataset, we highlight the contributions of our dataset as follows:




\textbf{Large scale.} Our dataset is large scale in terms of both the number of images and the number of categories.
  We collect images of \textbf{2388} different products (stock keeping units or SKUs), each containing 160 images on average, rendering a dataset with nearly \textbf{400k} images. 
  We also include the 10k original shelf images, with an average resolution of \textbf{3024 x 4032}.
  
\textbf{Realistic retail environment.} Unlike many of the previous datasets whose images taken under the laboratory environment or using web images, our data are all manually captured in multiple retail stores. The collection of our dataset matches the settings of realistic applications.

\textbf{Rich annotations.} Besides the individual SKU ID, we also provide multiple levels of annotations for each image.
The 2388 SKUs can be categorized into 7 meta categories based on their product types, or another 7 meta categories based or their shapes.
In addition, we provide the detailed attributes of each SKU, including the brand, flavor/type, and size, which allows the users to evaluate their algorithms on a customized fine-grained level. We will discuss this in detail in \secref{dataset}.

%% file: related.tex
\section{Related work}

\begin{figure}[t]
  \centering
    \includegraphics[width=1.0\columnwidth]{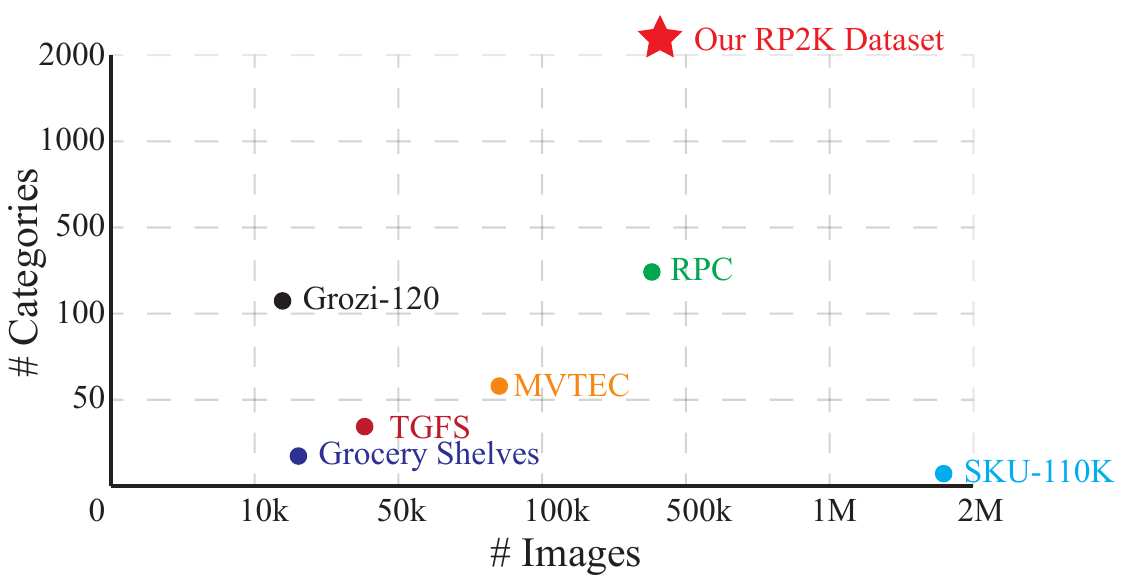}
    \caption{Comparing to other datasets, our dataset shows considerably larger number of categories, while maintaining a
    sufficient large amount of images.
  }\label{fig:cmp_dataset}
\end{figure}

\subsection{Fine-grained image classification}
Fine-grained image classification, the task of
recognizing images from multiple subordinate categories of a meta-category, has been extensively studied in computer vision research.
The challenging part of fine-grained classification is that objects belong to different subordinate categories may still look very similar. Thus, additional process is usually required for training a robust fine-grained classifier.
To benchmark the performance of fine-grained image classification algorithms, many datasets have been proposed in the past few years, including distinguishing different animal species~\cite{WelinderEtal2010,KhoslaYaoJayadevaprakashFeiFei_FGVC2011,berg2014birdsnap},
car models~\cite{KrauseStarkDengFei-Fei_3DRR2013} and clothes~\cite{liu2016deepfashion}.

Retail product recognition also lies in the field of fine-grained image classification, as products from the same brand but with different flavors may still look very similar even to human eyes, see \figref{example} for example.

\begin{figure*}[t]
  \centering
    \includegraphics[width=2.0\columnwidth]{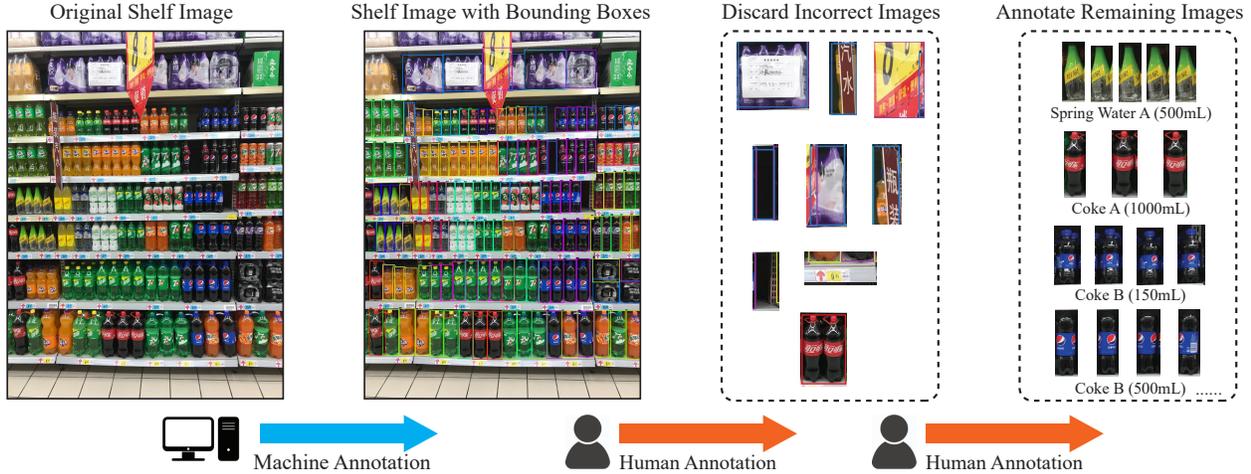}
    \caption{ Data collection pipeline of RP2K.  
    Our photo collectors were first distributed in over 500 different retail stores and collected over 10k high-resolution shelf images.
    Then we use a pre-trained detection model to extract the bounding boxes of potential objects of interests.
    After that, our human annotators discard the incorrect bounding boxes, including heavily occluded images and images that is not a valid retail product. The remaining images are annotated by the annotators.
  }\label{fig:pipeline}
\end{figure*}

\subsection{Retail product dataset} 
Even before deep learning dominated the majority tasks of computer vision, researchers have investigated recognition algorithms on retail products for decades.
A number of retail product datasets have also been proposed to support the research. 
For example, SOIL-47~\cite{soil-47} dataset contains 47 categories and 987 images which were acquired under different geometry of light sources. 
Grozi-120~\cite{grozi-120} was proposed for recognizing groceries in physical retail store. It contains 120 grocery product categories and 11,870 images.
Similarly, the supermarket product dataset~\cite{supermarket} focuses on fruit and vegetable classification, which contains 15 categories and 2,633 images.
There are many other similar datasets~\cite{george2014recognizing,varol2015toward,karlinsky2017fine,varol2015toward}, and we refer the reader to the survey~\cite{santra2019comprehensive}
for a comprehensive list of retail product datasets. 
Those early datasets usually have less than 20,000 images and may not suitable for today's data-demanding deep learning model. Here we review some recent datasets that are most relevant to ours.

\textbf{RPC dataset}~\cite{wei2019rpc} is a large-scale dataset proposed for automatic checkout.
It contains 200 categories and 83,739 images. Each image includes a different number of products (from 3 to 20) based on their clutter mode. Since bounding boxes and labels are provided for each object in each image,
it also yield 400k single-object images for object recognition task. However, the images were captured under controlled lighting and clean background, thus this dataset may not able to reflect the real-life scenario of product recognition on shelves.

\textbf{Take Goods from Shelves (TGFS)}~\cite{hao2019take} dataset is a object detection/recognition dataset for automatic checkout. Unlike RPC dataset captured in laboratory environment, TGFS dataset use images collected from self-service vending machines. Thus, the image distribution is closer to the checkout system in natural environment. Yet, it contains only 30K images from 24 fine-grained and 3 coarse classes, and the resolution of each image is 480 x 640, limiting the usage of this dataset.

\textbf{SKU-110K dataset}~\cite{goldman2019precise} is so far the largest retail image dataset in terms of the number of images. It contains more than 1M images from 11,762 images of store shelves. Since the main focus of the dataset is retail object detection in densely packed scenes, they only provide bounding boxes of each object in scene without further annotating the category of the bounding boxes.
Therefore, this dataset cannot be used for object recognition purpose.

\textbf{MVTEC}~\cite{follmann2018mvtec} is an instance-aware semantic segmentation dataset for retail products. It
provides 21,000 images of 60 object categories with pixelwise labels.
It can also be served as additional grocery relevant component to other semantic segmentation datasets~\cite{cordts2016cityscapes,everingham2015pascal,lin2014microsoft}.
Similar to RPC dataset, MVTEC is also captured in laboratory environment with controlled camera settings. The scale of this dataset in terms of images and categories is also relatively small. It may not be suitable for the task of object recognition in real store. 

Our dataset, in contrast to all previous dataset, designed particularly for product recognition on shelves.
Not only all images in our dataset are captured manually in real retail store (unlike many previous datasets use web images or capture images in lab environment),
we also provide so far the largest amount of SKUs with sufficient number of images for each SKU.
A comparison between our dataset and other dataset can be found in \figref{cmp_dataset}.

%% file: org.tex
\section{RP2K dataset}\label{sec:dataset}

\begin{figure}[b]
  \centering
    \includegraphics[width=1.0\columnwidth]{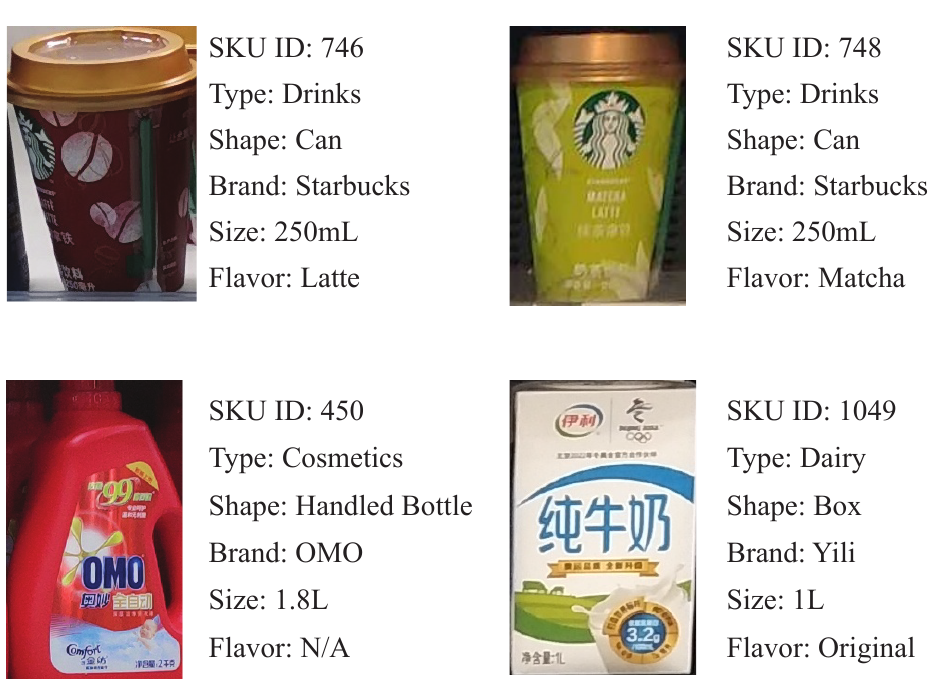}
    \caption{Sample data with attributes information.
  }\label{fig:attr}
\end{figure}

\paragraph{Organizations.}
Our dataset contains two components: the original shelf images and the individual object images cropped from the shelf images.
The shelf images are labeled with the shelf type, store ID, and a list of bounding boxes of objects of interest.
For each image cropped from its bounding box, we provide rich annotations include the SKU ID, product name, brand, product type, shape, size,
flavor/scent and the bounding box reference to its corresponding shelf image. 
\figref{attr} demonstrates some sample attributes of the object images. 
Note that some attributes may not be applicable to particular products.

We also provide meta category label for each object image in two different ways.
One is categorized by its product type, which reflects the placement of the products, i.e., products with the same type usually placed on the same or nearby shelf.
We include 7 meta categories by product types: {\fontfamily{lmss}\selectfont
dairies, liquors, beers, cosmetics, non-alcoholic drinks, tobacco} and {\fontfamily{lmss}\selectfont
seasonings.
}

\begin{figure}[t]
  \centering
    \includegraphics[width=1.0\columnwidth]{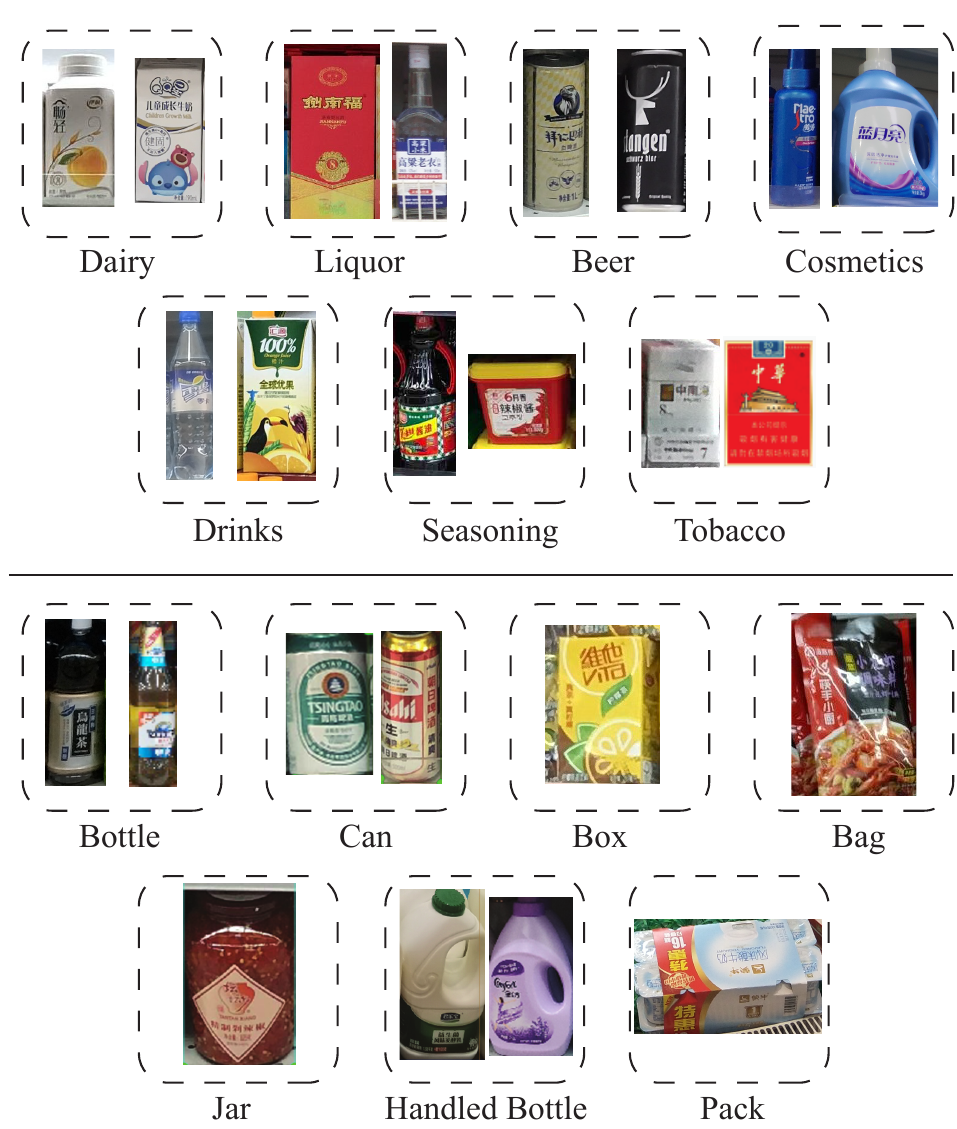}
    \caption{Sample images from our dataset with different meta-categorizations. \textbf{(top)} Categorized by product types. 
    \textbf{(bottom)} Categorized by product shapes.
  }\label{fig:sample}
\end{figure}

Another categorization method is by its product shape. We include 7 shapes, {\fontfamily{lmss}\selectfont
bottle, can, box, bag, jar, handled bottle} and {\fontfamily{lmss}\selectfont
pack
}, which covers all possible shapes that appeared in our dataset. These 7 shapes are also used in training our pre-annotation detector.
The sample images for different meta-categories are shown in \figref{sample}. 

Besides these two meta categorization method, our rich labels provide an option for the users to evaluate their algorithms on a customized fine-grained level.

For applications in retail product recognition, categorizing by product types is considered to be useful because a shelf image usually contains only one type of products. Thus, in our benchmark experiment (\secref{benchmark}) we focus on the product type meta category.


\paragraph{Data collections.}
The data collection and annotation pipeline is summarized in \figref{pipeline}.
The shelf images were collected in 10 cities from 500 different stores.
The data collectors were instructed to make sure the shelf is positioned in the center of the image, and each image should only contain one shelf.
The average resolution of shelf image is 3024 by 4032 pixels.
The image collectors use different (smartphone) cameras to capture the photos under natural in-store lighting environments, mimicking the realistic application scenarios.
We also make sure each individual object in the collected images is at least 80 by 80 pixels and clear enough to be recognized by human eyes.

Before we annotate the collected data, we first run a pre-annotation step to reduce the human workload.
We train a object detector on our auxiliary object detection dataset that can identify the 7 shapes we mentioned previously.
We use RetinaNet~\cite{lin2017focal} as the base structure of the object detector.
The object detector evaluate all collected shelf images, and propose bounding boxes for potential object of interests on the shelf images.

Then, we sent the pre-annotated shelf images to our human annotators. The annotators will first inspect the bounding boxes on the shelf images, i.e., removing all unwanted bounding boxes (e.g., overlapped box, occluded box and empty boxes) as well as adding missing boxes. After all bounding boxes have been validated, the annotator label each bounding boxes with its meta category, SKU ID and other attributes.



\paragraph{Statistics.}
We collect 10,385 high-resolution shelf images in total, with, on average, 37.1 objects in each images. The dataset contains in total 384,311 images of individual objects. 
Each individual object image represents a product from the 2388 SKUs. We split the train/test set by the ratio of 0.9/0.1.
The detailed statistics for different meta categories are represented in \tabref{stats}.




\begin{table*}[h]

  \centering

\begin{tabular}{ll|ccccc}
\bottomrule
&Meta Category & SKUs  & Train Imgs. & Test Imgs. & Total Imgs. & Imgs./SKU \\ \hline
\parbox[t]{0mm}{\multirow{6}{*}{\rotatebox[origin=c]{90}{By Product}}} 
&Dairies  & 323  & 78,288 & 8,867  & 87,155 &  269.8\\
&Liquors & 167  & 16,753 & 1,939  & 18,692  &  111.9\\
&Beers & 237  & 39,786  & 4,540 & 44,326 &    187.0  \\
&Cosmetics & 256  & 7,393  & 932  & 8,325  &   32.5\\
&Non-Alcohol Drinks & 333 & 29,241  & 3,405  &  32,646  &  98.0\\
&Seasonings &  560 & 137,082 & 15,479  & 152,561 & 272.4 \\
&Tobacco & 512 & 36,311 & 4,295 & 40,606 & 79.3\\
\hline


\parbox[t]{0mm}{\multirow{7}{*}{\rotatebox[origin=c]{90}{By Shape}}} &Bottle &852 &  164,939 & 18,327  & 183,266   & 215.1  \\
&Can  & 267  & 44,461  & 4,940  & 49,401  & 185.0 \\
&Box  & 411 & 27,347  & 3,039 &  30,386  & 73.9 \\
&Bag  & 198 & 15,350  &  1,705 &   17,055  &  86.1 \\
&Jar  & 246 & 13,913  &  2,657  & 16,570 &  67.35 \\
&Handled Bottle  &  251  & 54,895   &  6,099 & 60,994 &  243.0 \\
&Pack  &163  & 23,949 & 2,690 &  26,639  &  163.4 \\ \hline

&Total & \textbf{2,388} & \textbf{344,854} & \textbf{39,457} & \textbf{384,311} & \textbf{ 160.9}\\ 
 \toprule
\end{tabular}
  \caption{Statistics of our dataset.}
  \label{tab:stats}
\end{table*}



\paragraph{Visualizations.}
To understand the data distribution of our dataset qualitatively, we use UMAP ~\cite{2018arXivUMAP} to visualize our dataset.
Compare to other popular visualization techniques like t-SNE, UMAP is considered to be fast and scales better on high dimensional
data. We resize all our images to 32x32 before we send it to UMAP.

\begin{figure*}[t]
  \centering
    \includegraphics[width=1.8\columnwidth]{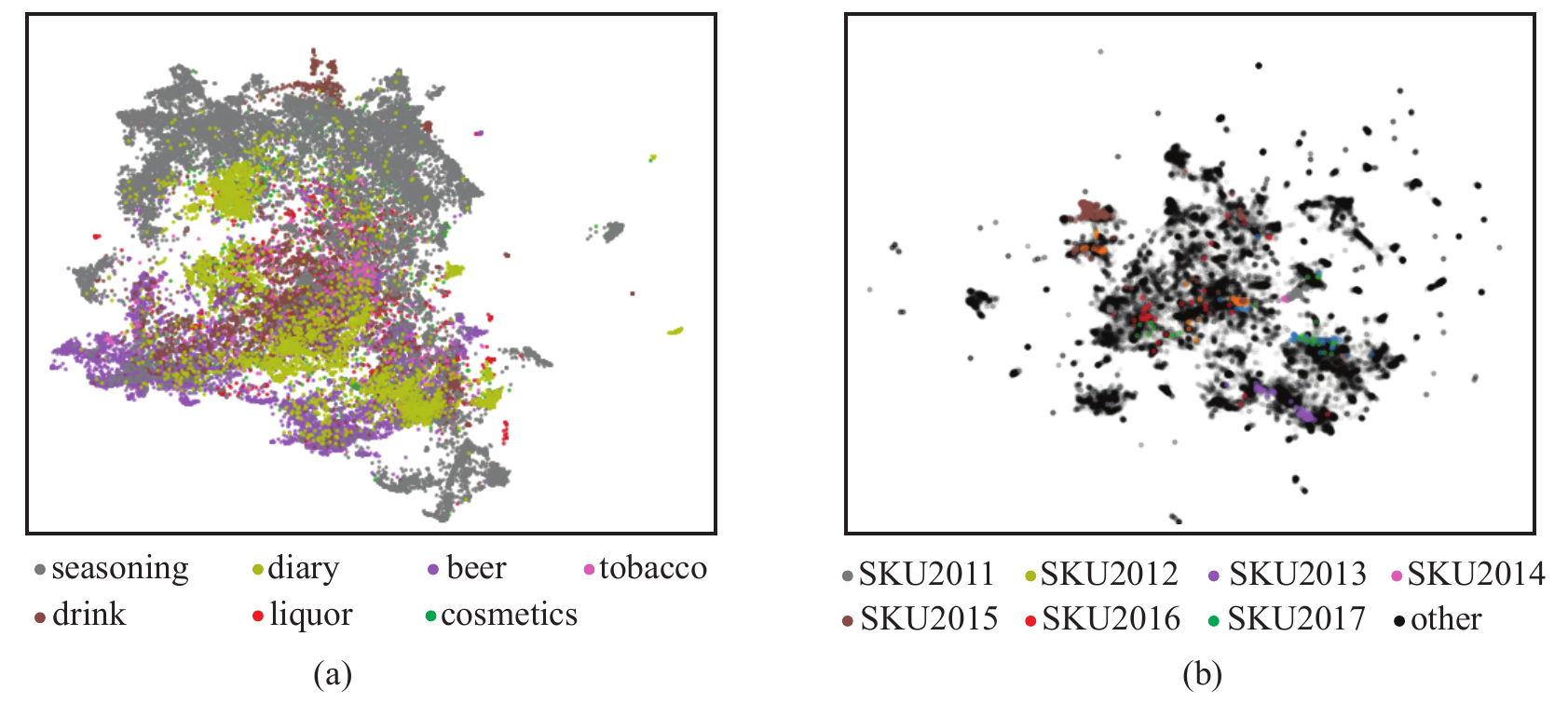}
    \caption{Visualization of data distribution via UMAP. \textbf{(a)} Visualization for all data categorized by its meta category. \textbf{(b)} Visualization for category {\fontfamily{lmss}\selectfont tobacco.} Here we only colorize SKUs with id from 2011 to 2017, while leaving all other points black.
  }\label{fig:vis}
\end{figure*}

\figref{vis} depict the visualization results. 
We found that most of the data are well separated when categorized by its shape. While some data points, especially those belong to category with relatively few images (e.g., cosmetics, liquors), are tend to spread across the data manifold.
In addition, when analyzing data in one meta category, for example, tobacco, we found that data with different SKU id do not separated well with each other. This reflects the coarse-to-fine nature of our dataset that the products from the different meta categories usually have distinguishable visual features, but products from the same meta category can look very similar even they are different products.

\paragraph{Auxiliary detection dataset.}
A full retail product recognition pipeline usually consists of two seperate parts---an object detector for locating the potential objects of interests and a classifier to recognize the object from the detected objects of interests.
To encourage the development of a full pipeline, we also provide an auxiliary object detection dataset that contains human-annotated bounding boxes of product in shelf images.
Since the object detection task has been addressed by many of the previous dataset~\cite{goldman2019precise,wei2019rpc,follmann2018mvtec}, we can further combine the data from previous work with our auxiliary detection dataset to build a larger scale detection dataset. Even without combining our data with other dataset, our pre-annotation practice indicate our auxiliary detection dataset is sufficient large for training a robust object detector.

For the detailed statistics, we include 95,800 bounding boxes from 1400 shelf images, labeled with 7 different shapes described in \figref{sample}. 
We want to emphasize that the object detection task for retail shelf has been specifically targeted by many other dataset,
thus we do not target our dataset as an object detection dataset.
Here we provide the detection dataset for completeness. 



%% file: baseline.tex
\section{Benchmarking our dataset}\label{sec:benchmark}
In this section, we provide our evaluations to benchmarking our RP2K dataset.
We demonstrate our results on the classification task. Yet, 
our dataset is not limited only to the classification task. We will provide a further discussion on other potential research problem of our dataset in \secref{future}.

\subsection{Evaluation of different classification algorithm}

We evaluate a number different classification methods on our dataset, including two standard classification networks (ResNet-34 ~\cite{he2016deep} and InceptionV3~\cite{szegedy2016rethinking}), as well as two state-of-the-art fine-grained classification methods (class-balanced loss~\cite{cui2019class} and API-Net~\cite{zhuang2020learning}). For both class-balanced loss (CBL) and API-Net we use ResNet-34 as its backbone network, and we use the default parameters and settings of their original implementation.
All experiments were done in Pytorch~\cite{paszke2019pytorch} framework and trained by Stochastic Gradient Descent (SGD) on a single NVIDIA Tesla V100 GPU.
We set the learning rate to 0.1 and the momentum to 0.9, while leaving other parameters to Pytorch default value. After every 50 epochs the learning rate will decay to 10\% of its original value, and we train each network 150 epochs. For each model, it generate one prediction from 2388 classes, and report both top-1 accuracies and top-5 accuracies among 7 product type meta categories. The results are summarized in \tabref{benchmark}.

\begin{table*}[t]

  \centering

\begin{tabular}{lc|cc|cc|cc|cc}
\bottomrule
\multirow{2}{*}{Meta Category}& \multirow{2}{*}{\# images} &\multicolumn{2}{c|}{ResNet-34} &  \multicolumn{2}{c|}{InceptionV3}& \multicolumn{2}{c|}{CBL} &  \multicolumn{2}{c}{API-Net} \\
 &  & Top-1 & Top-5 &  Top-1 & Top-5 & Top-1 & Top-5 &  Top-1 & Top-5 \\ \hline

Dairies & 87,155 & 93.68\% & 98.45\% & 93.38\% & 98.31\%    & 83.90\% & 94.26\%             & 54.30\%  &    80.80\%         \\
Liquors & 18,692 & 76.99\% & 96.44\% & 79.57\% & 97.31\%    & 61.88\% & 77.51\%             & 96.47\%  &    99.33\%         \\
Beers & 44,326& 96.76\% & 99.16\% & 95.19\% & 98.98\%       & 90.52\% & 95.85\%             & 97.79\%  &    99.60\%          \\
Cosmetics & 8,325 & 87.87\% & 93.71\% & 86.58\%  & 96.56\%  & 50.75\% & 65.23\%             & 93.88\%  &    99.03\%         \\
Non-Alcohol Drinks & 32,464 & 94.86\% & 98.88\% & 93.65\% & 98.53\%  &80.52\% & 90.57\%     & 96.91\%  &    99.38\%         \\
Seasonings & 152,561 & 97.96\% & 99.56\% & 97.28\% & 99.53\% &   93.44\%  & 97.64\%       & 98.71\%  &    99.86\%              \\ 
Tobacco & 40,606 & 91.89\% & 98.06\% & 91.73\% & 97.88\%    & 78.41\% & 85.44\%             & 93.15\%  &    98.76\%          \\ \hline
All & 384,311 & 95.18\% & \textbf{99.01\%} & 94.69\% & 98.96\%       & 87.66\% & 93.79\%             & \textbf{95.34\%}  &    98.65\%             \\

 \toprule
\end{tabular}
  \caption{Classification results for different categories. \# images indicates the number of total images available in the corresponding meta-category. The simple ResNet-34 network achieves a very similar performance as API-Net, which is the state-of-the-art method on many fine-grained classification task.}
  \label{tab:benchmark}
\end{table*}

\paragraph{Results.}
Although CBL and API-Net reportedly achieves the state-of-the-art results on multiple fine-grained classification dataset like CUB-200-2011 and Stanford Cars, they did not surpass even a simple ResNet-34 network on RP2K dataset in terms of the recognition accuracy.
In addition, as we mentioned previously, a classification system with 95\% accuracy may considered to be very good from the computer vision perspective, but not enough for industrial level retail application.
This indicate that there are still some room for improving the recognition accuracy.

Besides, we found that although ResNet-34 achieves an overall accuracy 95\%, there are some categories like liquors and cosmetics have much lower accuracy than average. This is because 1) the number of training data of those categories are relatively smaller than others; 2) the appearance of products in those categories usually look very similar.

\subsection{Evaluation of different training scheme}

Since images in our dataset are captured under diverse view angles and lighting conditions, using data augmentation during training presumably would increase the classification performance. 
In addition, we are also interested in evaluating the performance between using a (ImageNet) pre-trained model and training from scratch.

Thus, we further evaluate the ResNet-34 using 4 different training schemes:
training from scratch, pre-training, training from scratch with data augmentation, and pre-training with data augmentation.
For all training scheme we use the SGD optimizer.


To mimic the real scenario environment, we use the following data augmentation scheme in our training:
\begin{itemize}
  \setlength\itemsep{-0.8mm}
	\item Adding a constant border with a width randomly chosen between 0 and 30px.
	\item Cropping randomly by up to 10px.
	\item Applying a randomly-parameterized perspective transformation.
	\item Darkening/Brightening the image ramdomly by up to 20\%.
\end{itemize}
The results of evaluating 4 training protocols are shown in \tabref{benchmark2}. 

\begin{table}[t]

  \centering

\begin{tabular}{l|cc}
\bottomrule

Training Method & Top-1 Acc. & Top-5 Acc.\\ \hline
Scratch & 95.18\% & 99.01\%\\
Pre-train & 95.54\% & 99.04\%\\
Scratch + Augmentation &  90.41\%  & 94.74\%\\
Pre-train + Augmentation & 90.89\% & 95.01\% \\

 \toprule
\end{tabular}
  \caption{Evaluation of different training protocols. We found that adding data augmentation does not necessarily increase the
  recognition performance. We hypothesize it is due to our dataset already including lighting variance and camera distortion, so that further augmenting data does not make our dataset closer to the real image.}
  \label{tab:benchmark2}
  \vspace{-6mm}
\end{table}







\begin{figure}[h]
  \centering
    \includegraphics[width=1.0\columnwidth]{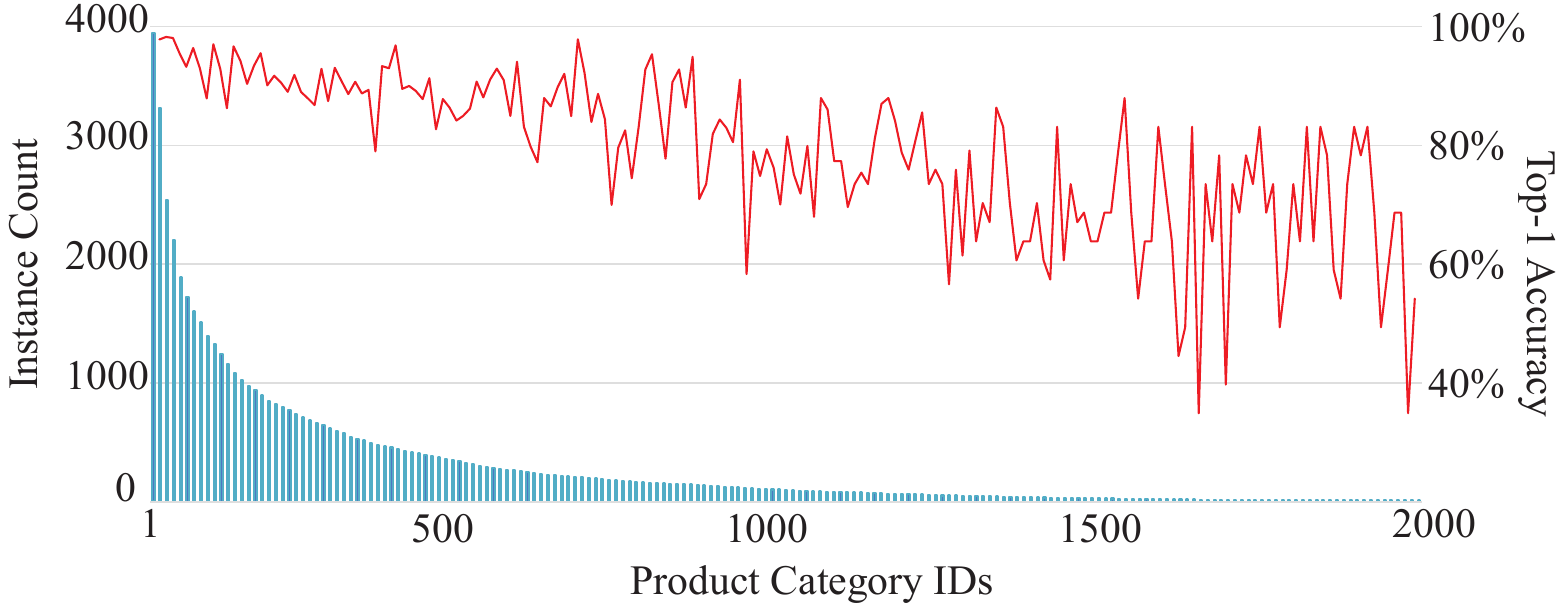}
    \caption{Long tail problem in fine-grained recognition. With the decreased number of available images, the recognition accuracy tends to decrease.
  }\label{fig:longtail}
\end{figure}

Besides reporting classification accuracy for each meta category, we also plot the relationship between the number of instance image and classification accuracy for each individual instance.
To analyze this relationship, we first sort our entire 2388 product by its instance count, and then group each 10 nearby products.
Then we calculate the average top-1 classification accuracy for each group.
As shown in \figref{longtail}, the prediction accuracy is decreasing with the number of instances available in the dataset, and the variance of the accuracy increased.
This suggest that future research should focus on improving the prediction accuracy for the "long tail" part, since total accuracy could be high due to instances with abundant training data usually have high prediction performance while contribute to the statistics more.
In contrast, the long tail part may still have low accuracies. 

\subsection{Auxiliary Detection Dataset} \label{sec:detection}
Here we provide the evaluation on the auxiliary detection dataset.
We split the 95,800 bounding boxes to 80,000 training boxes and 15,800 test boxes.
We use the standard RetinaNet~\cite{lin2017focal} to evaluate the detection performance.
We report the Average Precision (AP) for each shape, as well as mean Average Precision (mAP) for the entire test set, with different IoU threshold, the results are summarized in \tabref{detection}.



\begin{table*}[h]

  \centering

\begin{tabular}{l|cccccccc}
\bottomrule
Shape & AP(0.5) & AP(0.55) & AP(0.6) & AP(0.65) & AP(0.7) & AP(0.75) & AP(0.8) \\ \hline
Box &  0.3485&  0.3475 &   0.344    & 0.3372 & 0.3195 &  0.3003 & 0.2670          \\
Can &   0.6886& 0.6868 &    0.6837  & 0.6811& 0.6766  &  0.6650 & 0.6314\\
Bottle & 0.7525& 0.7509 &   0.7487  & 0.7448& 0.7365  &  0.7137 & 0.6545\\
Jar &   0.2620&  0.2559 &   0.235   & 0.2302& 0.2191  &  0.1921 & 0.1748  \\
Handled Bottle & 0.4919& 0.4896  &  0.4658 & 0.4505 & 0.4014 & 0.3804 & 0.3059\\
Bag & 0.3449 &  0.338 & 0.3273 & 0.3151& 0.2827 & 0.2518 & 0.1901          \\
Pack & 0.4643 & 0.4634 & 0.4634 & 0.4634&  0.4634 & 0.4494 & 0.3789\\ \hline
mAP for All & 0.6186&  0.6164 & 0.6121 & 0.6069  & 0.5962 & 0.5739 & 0.3994 \\

 \toprule
\end{tabular}
  \caption{Performance of object detection. AP($x$) indicates AP with IoU threshold=$x$.}
  \label{tab:detection}
\end{table*}

\section{Other potential research problems}\label{sec:future}
Besides the classification and detection tasks, there are many other potential research problems that can be explored on our dataset.
We list a few of them in this section.
\paragraph{Adversarial attacks and defenses.}
Adversarial attacks refer to adding deliberately crafted, imperceptible noise to natural images, aiming to mislead the network's decision entirely~\cite{goodfellow2014explaining}. Adversarial attacks pose serious
threats to numerous machine learning applications, from autonomous driving to face recognition authorization.

To launch adversarial attacks, the adversary needs to solve the optimization problem:

\begin{align*}
\max_{\delta x} L(f(x+\delta x), y), 
\end{align*}
where $f$ is the targeted neural network, $x$ is the input data, $\delta x$ is the applied perturbation, $y$ is the ground truth label for $x$ and $L$ is the surrogate loss function for training $f$.

To solve this non-convex optimization problem, one simple yet effective way is to use Projected Gradient Descent (PGD)~\cite{madry2018towards} via the following iteration:

\begin{align*}
    \delta x_{n+1} = \Pi_{\delta x \in \Delta_{\epsilon}} \delta x_n + \alpha \text{sign} \nabla_{\delta x}L(f(x+\delta x_n), y),
\end{align*}
where, $\alpha$ is the step size, $\Delta_{\epsilon}$ is the allowed perturbation range and $\Pi$ is a projection operator that projects $\delta x$ to its constraint space $\Delta_{\epsilon}$. Here we consider the $\ell_{\infty}$ threat model, i.e., $\delta x \in \Delta_{\epsilon}$ if and only if $||\delta x||_{\infty}<\epsilon$.

To defend such adversarial examples, the most recognized way is through adversarial training~\cite{madry2018towards}: training on the adversarial examples generated at each training epoch. This method has been shown to be effective on many popular datasets.
Motivated by this, a series of adversarial attacks and defenses have been proposed, which advance the understanding of adversarial examples~\cite{cohen2019certified,shafahi2019adversarial,Xiao_2020_CVPR,tramer2020adaptive}.

However, most of the existing defense methods are evaluated on standard benchmarks such as MNIST, CIFAR-10 or ImageNet.
In those datasets, the visual difference between images from different classes are often large.
In contrast, due to the fine-grained characteristics, two images from different classes in our dataset could hold very similar visual features. 
Besides, the total number of categories in our dataset (2388) is also considerably higher than traditional benchmarks like CIFAR-10 (10) or ImageNet (1000). 
These two factors could yield a much more challenging problem in the adversarial defense task.

To better understand the property of our dataset under the adversarial learning perspective, we run experiments to evaluate the accuracy of models (with or without adversarial training) under adversarial attacks.
Here we use standard training protocol~\cite{madry2018towards} for adversarial training and PGD to generate adversarial examples.
We evaluate it under $\ell_{\infty}$ threat model with perturbation size $\epsilon=4/255$ and $8/255$.

As shown in \tabref{adv}, adversarial training not only drastically decreases the accuracy on standard test images, but also shows little effect on defending adversarial examples.
Given the failure of the most popular defense method, we believe our dataset could also serve as an alternate benchmark for evaluating adversarial defense and attack algorithms.

\begin{table*}[t]

  \centering

\begin{tabular}{l|ccc}
\bottomrule


Model & Clean Acc. & Adv. Acc. ($\epsilon=4/255$) & Adv. Acc. ($\epsilon=8/255$) \\ \hline
Regular & 95.18\% & 0.00\% & 0.00\% \\
Adv. Train ($\epsilon=4/255$) & 40.14\% & 12.35\% & 0.00\% \\
Adv. Train ($\epsilon=8/255$) & 20.90\% & 11.74\% & 5.26\%\\

 \toprule
\end{tabular}
  \caption{Accuracies on RP2K under Adversarial attacks. We use 3 different training scheme to train a ResNet-34 base model: regular training, adversarial training with $\epsilon=4/255$ and adversarial training with $\epsilon=8/255$. And we evaluate their clean accuracy, adversarial accuracy with $\epsilon=4/255$ and adversarial accuracy with $\epsilon=8/255$. When evaluating on standard test images, models with adversarial training shows drastically decreased accuracy compare to regular trained model.
  When evaluating on adversarial test images, regular model can always be fooled by adversarial attacks, while adversarial trained models only show little effectiveness for defending adversarial examples.}
  \label{tab:adv}
  \vspace{-6mm}
\end{table*}

\paragraph{Generative models on structured images.}
Image synthesis has achieved remarkable progress in recent years with the emergence of various generative models~\cite{radford2015unsupervised,arjovsky2017wasserstein,xiao2018bourgan,mustafa2019cosmogan,karras2019style,kotovenko2019content}.
The state-of-the-art generative models are capable of generating realistic high-resolution images of many distinct objects and scenes.

Despite the success in generating natural images, generating images with structured layouts remains challenging, as pointed out by many recent work~\cite{qi2019ke,hinz2018generating,Sun_2019_ICCV}.
Our original shelf image dataset could be a practical dataset for evaluating generative models on structured image synthesis. 
The bounding box combined with SKU labels in our dataset provides the ground truth of semantic layout information.
Once robust generative models are developed, it could also facilitate to pave a way for generating more data for the tasks of shelf objects detection and recognition.


\paragraph{Few-shot learning.}
The ability to learn from a few examples remains a challenge for modern machine learning systems. This problem has received significant attention from the machine learning community~\cite{finn2017model,ren2018meta,snell2017prototypical,duan2017one,garcia2018fewshot}.
According to \figref{longtail}, the long-tail effect of our characteristics provides more than 100 classes with the number of instance images less than 30. Thus, our dataset could also serve the purpose of few-shot learning algorithm evaluation.
Moreover, the large number of categories reside in our dataset enables a broader range of choices to evaluate the algorithm.
Besides the research value of few-shot learning in our dataset, few-shot learning is also important in the retail industry since some products may only have few placed in a retail store.




\section{Discussion: applications in retail}\label{sec:app}

A robust retail object recognition system is desired in a myriad of applications in AI-powered retail store. 
It is foreseeable that our dataset can be served as a starting point for any retail applications using shelf images. Future applications based on retail object recognition system include but not limit to:
\paragraph{Auditing product placement.} Auditing shelf management using object recognition system plays an essential role in understanding the shelf conditions. On the one hand, costumers may mistakenly place products they don’t want to buy on the wrong shelf. Identifying the wrong placement requires a lot of human resources in traditional retail store, but it can be immediately
detected by computer system. On the other hand, analysis show that customer make important buying decisions based on the placement of store shelves~\cite{law2004product}, and retailers invest a lot to create ideal planograms as their selling strategy. It is crucial to make sure that the products are placed in their desired way.
\paragraph{Detecting empty shelf.} Empty shelf is an indicator of whether a product is out-of-stock. In many cases, the retailers have more available stocks in their warehouse even their store shelves show out-of-stock. Before the retailer refill the stocks, it even takes a certain amount of time for the retailer to realize the product is out-of-stock. With a robust recognition deployed in the store, the store can quickly send notification of which product is out-of-stock and tracking the selling activities in
real-time.
\paragraph{Image-based product retrieval.} Not only for the retailer, the costumer could also benefit from a
well-established in-store object recognition system. Imagine you want to search information for a
particular product in-store, instead of Googling the product name on web, you can take a photo of
the product and the information will immediately prompt out. Moreover, based on the image, your
location information could also be retrieved, rendering a image-based navigation system. This could
increase the accessibility of the retail store.

%% file: conclusion.tex
\section{Conclusion}
We introduce a new fine-grained retail recognition dataset, RP2K, which contains a large number of retail product images from 2388 different products. 

Our dataset is inspired by the task of retail product recognition on store shelves, which has tremendous applications on AI-powered retail industry---image-based product retrieval, empty shelf detection, and sales activity tracking, to name a few.
As a fine-grained classification dataset, our dataset has the so far largest amount of categories, while maintaining a sufficient amount of images.
In addition to the pre-defined meta-categories, we also provide rich attributes information that allows the user to adjust the fine-grained level for their evaluations.

To bridge the gap between research and real-life applications, our data are all collected in natural retail store environments.
Our experiments show that, on our dataset, even the state-of-the-art fine-grained classification method cannot achieve significantly better results than a simple ResNet baseline, indicating that there still exist large room for improvement.

Besides object recognition, our dataset could also be used for other important computer vision tasks such as adversarial learning and generative model.
We believe our dataset could further progress the revolution that is already occurring in the retail industry, and lead to a step toward building a fully autonomous retail store.